**Title:**
How Robot Dogs See the Unseeable: Improving Visual Interpretability via Peering for Exploratory Robots

**Authors:**
Oliver Bimber[1*], Karl Dietrich von Ellenrieder[2], Michael Haller[3], Rakesh John Amala Arokia Nathan[1], Gianni Lunardi[2], Mohamed Youssef[1], Marco Camurri[4], Santos Miguel Orozco Soto[2], Jeremy E. Niven[5]

**Affiliations:**
[1]Department of Computer Science, Johannes Kepler University Linz, Austria
[2]Field Robotics Laboratory South Tyrol, Free University of Bozen-Bolzano, Italy
[3]Media Interaction Lab, Free University of Bozen-Bolzano, Italy
[4]Department of Industrial Engineering, University of Trento, Italy
[5]School of Life Sciences, University of Sussex, UK

*Corresponding author. Email: oliver.bimber@jku.at

**Abstract:** In vegetated environments, such as forests, exploratory robots play a vital role in navigating complex, cluttered environments where human access is limited and traditional equipment struggles. Visual occlusion from obstacles, such as foliage, can severely obstruct a robot's sensors, impairing scene understanding. We show that "peering", a characteristic side-to-side movement used by insects to overcome their visual limitations, can also allow robots to markedly improve visual reasoning under partial occlusion. This is accomplished by applying core signal processing principles, specifically optical synthetic aperture sensing, together with the vision reasoning capabilities of modern large multimodal models. Peering enables real-time, high-resolution, and wavelength-independent perception, which is crucial for vision-based scene understanding across a wide range of applications. The approach is low-cost and immediately deployable on any camera-equipped robot. We investigated different peering motions and occlusion masking strategies, demonstrating that, unlike peering, state-of-the-art multi-view 3D vision techniques fail in these conditions due to their high susceptibility to occlusion. Our experiments were carried out on an industrial-grade quadrupedal robot. However, the ability to peer is not limited to such platforms, but potentially also applicable to bipedal, hexapod, wheeled, or crawling platforms. Robots that can effectively see through partial occlusion will gain superior perception abilities – including enhanced scene understanding, situational awareness, camouflage breaking, and advanced navigation in complex environments.

**One Sentence Summary:** By mimicking how insects peer side-to-side, robots can now see through clutter in real-time.

**Main Text:**

## INTRODUCTION

Peering is the characteristic side-to-side body movement animals use to judge distance before jumping or striking (Fig. 1, top-left). Animals with limited, short range stereovision exploit motion parallax with movements that create relative motion between nearby and distant objects. The peering of insects, such as praying mantises (1), horse-head grasshoppers (2), locusts (3), or crickets (4), is well studied. In locusts, for instance, leg movements pivot the body around the abdomen, producing a lateral head displacement that is compensated by an opposite head rotation to stabilize eye orientation (3). Such behaviour patterns increase motion parallax. Whether peering serves functions in occluded environments (for instance, to see prey through foliage) remains unexplored.

Peering quadrupedal robots can see through partially occluding ground vegetation and other obstacles (Fig. 1, top-right). Potential applications are manifold and range from surveillance, terrain exploration, inspection, search-and-rescue, and more. This work establishes a connection between peering in visual perception and synthetic aperture sensing in optical imaging. We show that this principle improves scene understanding not only in animals but also in robots.

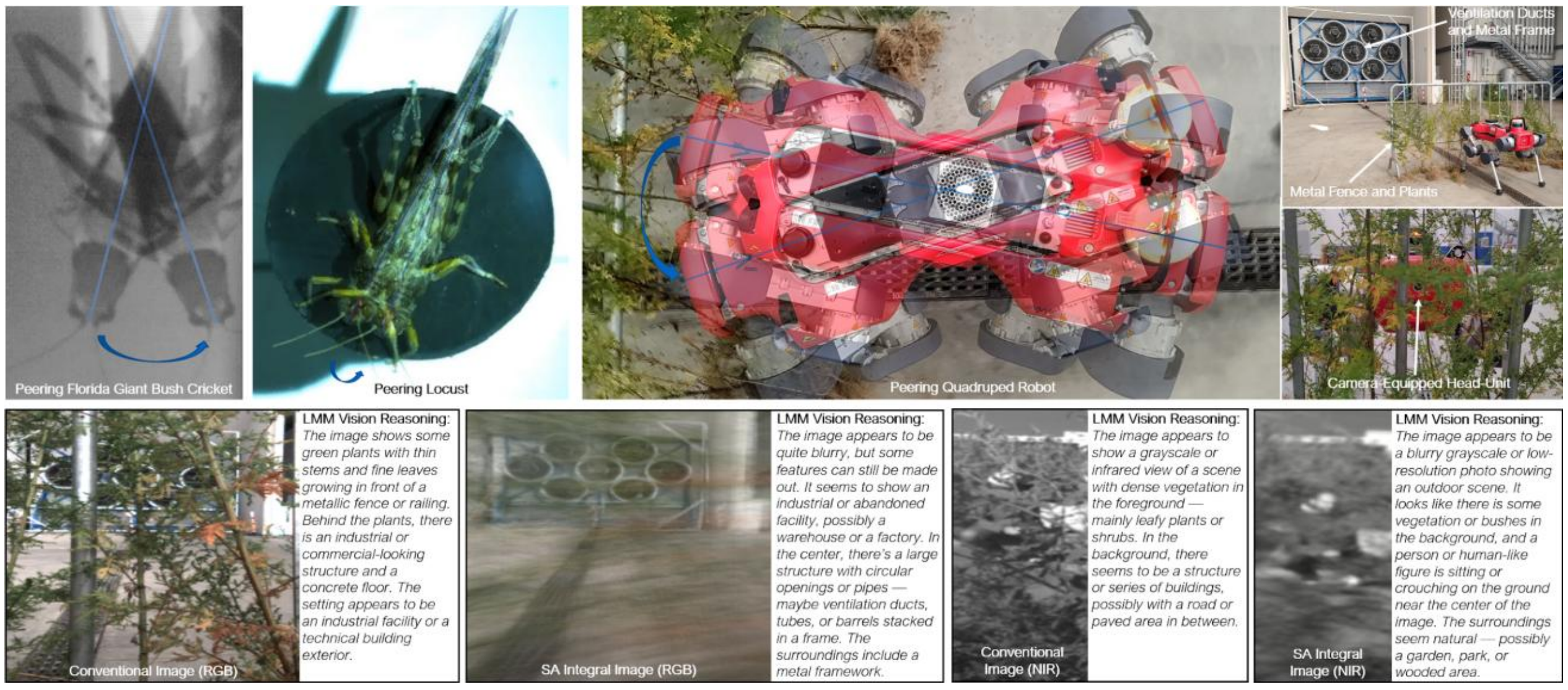

**Fig. 1. Animal and robotic peering enhance scene understanding under limited vision.** Using side-to-side peering motions, insects like katydids (e.g., bush crickets) and locusts leverage motion parallax to perceive depth beyond what stereoscopic vision provides. Similarly, robots with limited aperture optics (e.g., the ANYbotics ANYmal) can use this strategy to suppress partial occlusion. While sophisticated vision reasoning of large multimodal models (here, ChatGPT-5.0) struggles to interpret occluded scenes when being asked to interpret visible content in conventional images, these models show improved understanding once occlusion is suppressed through peering in synthetic aperture (SA) integral imagery. Robot peering suppresses occlusion in real-time and is wavelength-independent, as demonstrated in the visible (RGB) and near-infrared (NIR) examples above. For the RGB example, a peering motion (=SA) of approximately 11cm was used to sample and integrate 300 images captured with a wide-angle lens. For the NIR example, the peering motion was approx. 7cm and integrated 27 images captured with a standard lens. The physical aperture of both cameras was 0.965mm (f/2). See Movie S1 in *Supplementary Materials*.

Bio-inspired robotics leverages principles observed in biological organisms to develop robotic systems that exhibit enhanced adaptability, resilience, and efficiency. This paradigm provides a rigorous foundation for engineering capabilities that emulate functional traits found in nature. For example, bio-inspired locomotion strategies enable robots to operate reliably in environments that are unstructured, harsh, unstable, deformable, or otherwise unsuitable for conventional robotic platforms (5-29). Likewise, bio-inspired perception draws on sensory and behavioral mechanisms evolved in diverse species to facilitate safe, efficient, and context-aware interaction within complex habitats (30-52). With active perception (53-62) a robot actively moves or emits signals to gather better information. Instead of passively observing, the robot adjusts its sensing to reduce uncertainty, improving perception in challenging environments. Collectively, these perspectives illustrate how biological principles inform advances in enhancing robotic perception and navigation. Furthermore, research inspired by vision in humans, insects, birds, and other animals has advanced depth estimation, obstacle avoidance, and motion perception in dynamic environments. Human-vision-inspired robotics studies have contributed to near-field 3D perception in humanoid robots. A biologically grounded model (38) reproduced compensatory head–eye coordination during rotations, producing reliable parallax for scene segmentation and egocentric distance. Building on this, other systems (39-41) demonstrated that fixational head movements generate usable motion parallax for depth inference, inspiring monocular depth estimation algorithms that integrate ego-motion and visual parallax cues.

Insect vision studies highlight optical flow as a key cue for navigation and collision avoidance. A fully neuromorphic, closed-loop robotic system modeled after flies and bees demonstrated efficient obstacle avoidance and gap detection by steering toward regions of low apparent motion through an inverse winner-take-all network (34). Another insect-inspired approach (35) integrated spatial pooling of optical flow with acceleration cues to estimate ground velocity and distance, addressing the scale ambiguity of monocular vision. The work in (51) reports a fully wireless, power-autonomous, mechanically steerable vision system that imitates insect head motion, successfully mounting it on live beetles and insect-scale robots. Active vision behaviors in insects and birds have also inspired depth estimation and 3D reconstruction. The work in (36) modeled mantis peering (lateral motion) and pigeon bobbing (axial motion) for static and dynamic depth perception, respectively. Similarly, the "mantis head camera" (37) replicated mantis peering to achieve accurate near-field depth estimation via motion parallax from optical flow.

Bird vision offers powerful inspiration for robotics, particularly in tasks requiring precise perception, active sensing, and agile navigation. Barn owl kinematics (42) inspired the development of a biomimetic active-vision robot that employs owl-like side-to-side peering trajectories, resulting in an improved 3D reconstruction accuracy compared with non-biomimetic scans. Similarly, (52) presents a bio-inspired active sense and avoid (ASAA) system for flying robots that uses a single mechanically steerable stereo camera –mimicking the owl's flexible neck and head– to overcome the narrow field-of-view limitation and safely navigate environments with dynamic obstacles.

All prior works on bio-inspired robot vision exhibit a critical limitation: their effectiveness substantially diminishes in the presence of occlusions. Overcoming this limitation is central to our contribution. This is achieved through the application of core signal processing principles, namely optical synthetic aperture sensing.

In classical optical imaging, cameras integrate light through the aperture of their lenses. The depth of field is the distance range, measured along the optical axis, on either side of the focal plane within which objects will appear acceptably sharp in an image. Typical robot cameras have small (e.g., less than a millimeter wide) optical apertures and are fixed focus, such that the whole scene appears sharp without the need to adjust the focus for image processing tasks (e.g., visual odometry). However, this design choice also carries a fundamental limitation: their large depth of field renders both near and far objects in sharp focus. Consequently, any obstacle in the foreground can obscure objects of interest in the background (Fig. 2). Synthetic aperture sensing (63) is an advanced signal processing technique where a moving sensor (like a radar antenna or optical camera) collects a series of measurements over time and location. These measurements are then computationally combined to synthesize the effect of a single, much larger aperture.

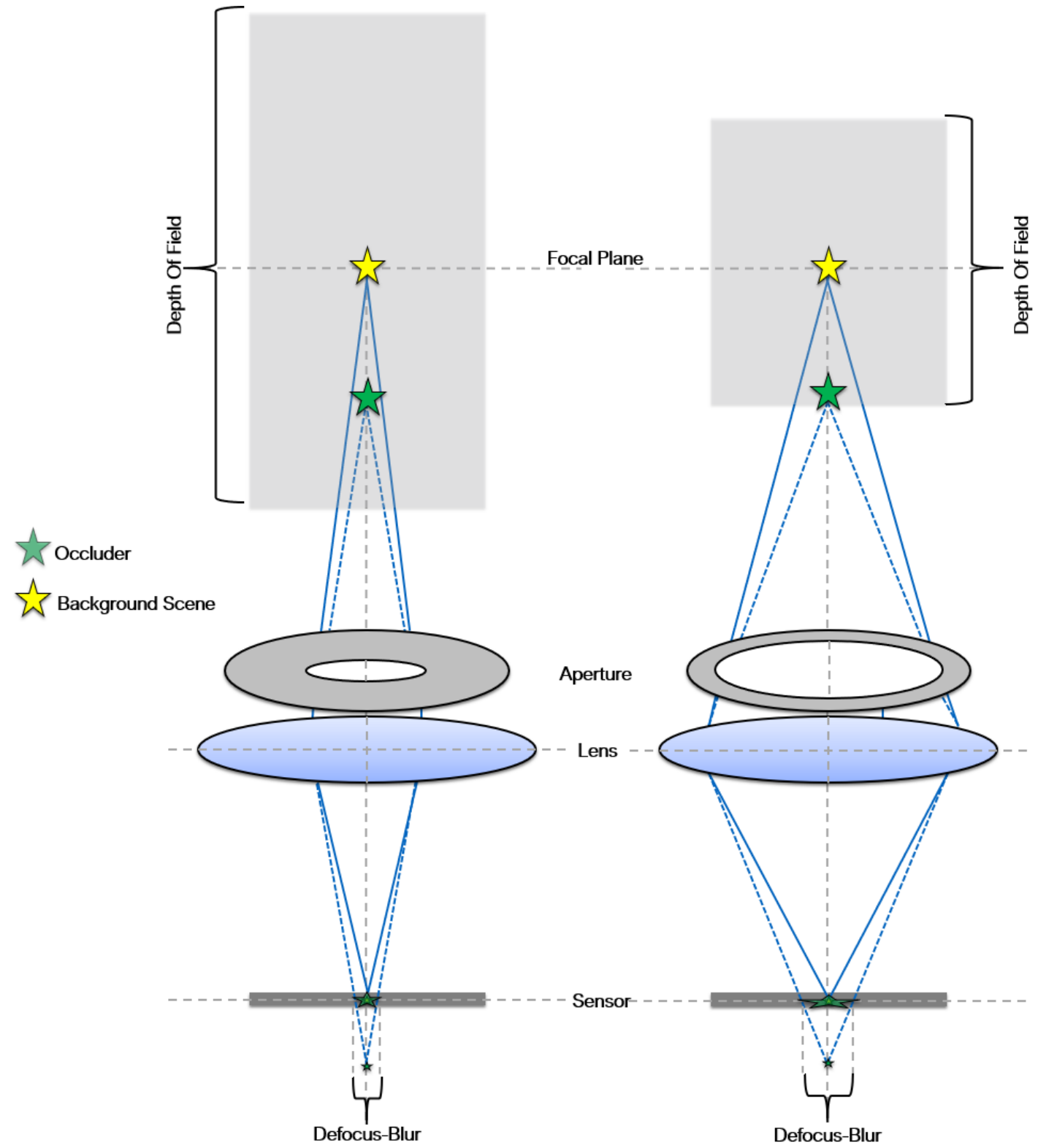

**Fig. 2. Depth of field decreases as the aperture widens.** A narrow aperture (left) produces minimal defocus blur from out-of-focus objects, resulting in a wide depth of field where both foreground and background appear sharp. Conversely, a wide aperture (right) admits more light but shrinks the depth of field, blurring anything outside the focal plane more. However, conventional camera optics limit the maximum aperture size based on the lens and sensor dimensions. Consequently, robots often use fixed focal-length cameras with sub-millimeter apertures, which keep the entire scene in sharp focus, making the background scene and occluders in the foreground appear equally defined.

In case of optical synthetic aperture sensing, images recorded at spatially varying poses are computationally integrated to produce an image with the extremely shallow depth of field that a physical lens of the synthetic aperture's (SA) size would create. As shown later, a synthetic focal surface, such as a plane, cylinder, sphere, or other more complex topologies can now be aligned to focus on scenery beyond the occluding foreground. The optical signal of occluders is then lost in the defocus blur, while the optical signal of the focused background is amplified. The computational process requires only the projection and averaging of the captured images onto the synthetic focal surface and is efficient enough for real-time operation on mobile processors. It is also wavelength-independent, delivering robust performance across all spectral bands –including visible (RGB), thermal (FIR), near-infrared (NIR), RedEdge (REG), and low-light spectra – making it versatile across numerous applications. The efficacy of synthetic aperture sensing for occlusion removal is highest at 50% occlusion (64), but degrades with increasing occlusion density. This principle was initially explored using fixed, structured camera arrays (65). However, unlike peering, camera arrays severely undersample the SA signal. They also introduce large, complex payloads, demand prohibitively high bandwidth to transmit multiple parallel video streams, and lack the ability to adapt sampling in response to local occlusions. Subsequent work extended the principle to enable extremely wide, unstructured SA sampling using drones and drone swarms (66). These systems have been applied to tasks including search and rescue (67, 68), wildlife observation (69), wildfire detection (70), and forest ecology (71). Research has further demonstrated that removing occlusions improves fundamental scene understanding tasks such as image classification (72).

## RESULTS

During peering, the robot's camera-equipped head describes a several centimeter wide SA in which hundreds of images and corresponding poses (computed from the robot's internal filter using inertial and kinematics information) are captured. The images can be recorded across arbitrary wavelengths to support a wide range of applications. They can be processed and integrated computationally in real time to suppress partial occlusion, thereby enhancing visual reasoning (Fig. 1, bottom). In the following section, we present results for different peering motions and occlusion masking strategies. Our findings demonstrate that peering is able to successfully recover objects in the background, while state-of-the-art 3D vision techniques fail under the same conditions.

### Why multi-view 3D vision fails

Synthetic aperture sensing does not reconstruct depth but instead suppresses partial occlusion in 2D images. An alternative to processing the recorded image samples for visual scene understanding is multi-view 3D vision. The experiment in Fig. 3 shows that current model-based and deep-learning techniques fail even under relatively sparse occlusion.

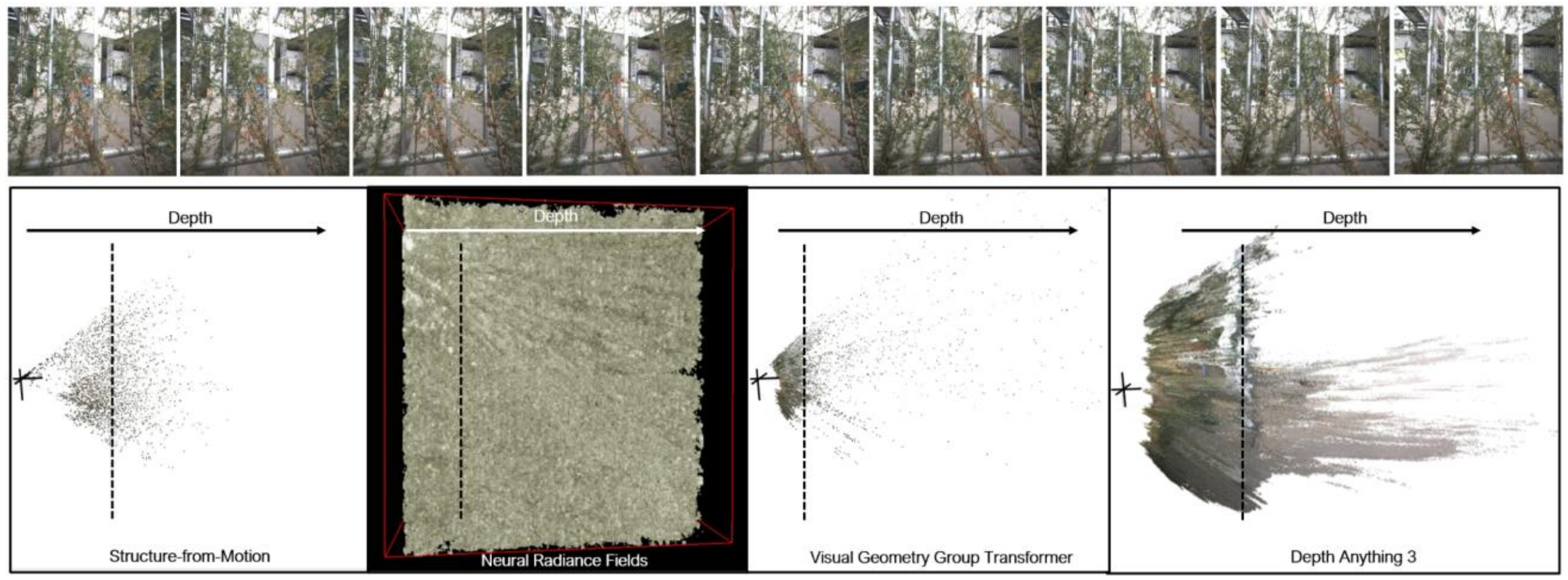

**Fig. 3. Multi-view 3D vision in the presence of partial occlusions.** The above example illustrates nine samples from the 300 wide-angle RGB images captured during the 11cm peering motion shown in Fig. 1, and alongside point-cloud and volume reconstruction results from state-of-the-art multi-view 3D vision methods. Coordinate frames mark the SA center. The dashed lines indicate the approximate distance to the nearby metal frame visible in the individual images. The occluded ventilation ducts and supporting structure in the background are not reconstructed.

We processed the 300 images from the wide-angle RGB peering experiment (Fig. 1) with several state-of-the-art 3D vision approaches: Structure-from-Motion, *SfM* (73); Neural Radiance Fields, *NeRF* (74); Visual Geometry Group Transformers, *VGGT* (75); and the dual-encoder feature-fusion network of Depth Anything v3, *DA3* (76).

The *SfM*, *VGGT*, and *DA3* methods reconstructed point clouds consisting primarily of the foreground occluders. The NeRF model produced a volume of largely incoherent noise. In all cases, the occluded background was not recovered due to random occlusion of image features across the multi-view input. This demonstrates that multi-view 3D vision techniques, which rely on consistent features across images, are highly susceptible to occlusion. They fail when these features are randomly hidden.

Furthermore, none of these approaches are capable of real-time processing, all require relatively long computational times. While *SfM* and *NeRF* require hours (5-6 hours in our experiment), *VGGT* and *DA3* need minutes (7-15 minutes in our experiment) of computation even on high-end GPUs.

The corresponding peering results shown in Fig. 1 were computed in milliseconds (17 milliseconds in our experiment) and enable high-resolution perception, which is essential for vision-based scene understanding.

**Synthetic aperture sensing via peering motion**

Occlusion suppression with peering is low-cost, wavelength independent, and easily deployable on any camera-equipped robot. The robot's peering motion allows it to capture hundreds of images using a conventional narrow-aperture camera. The trajectory of this motion defines a larger SA (Fig. 4, left). To synthesize this aperture, the images are projected onto a synthetic focal surface and averaged (Fig. 4, right), a process that requires precise camera pose data from the robot's sensors. The resulting SA, which can be several centimeters wide, creates an extremely shallow depth-of-field SA integral image. Only the regions near the synthetic focal surface remain in sharp focus, while all other areas are rapidly blurred.

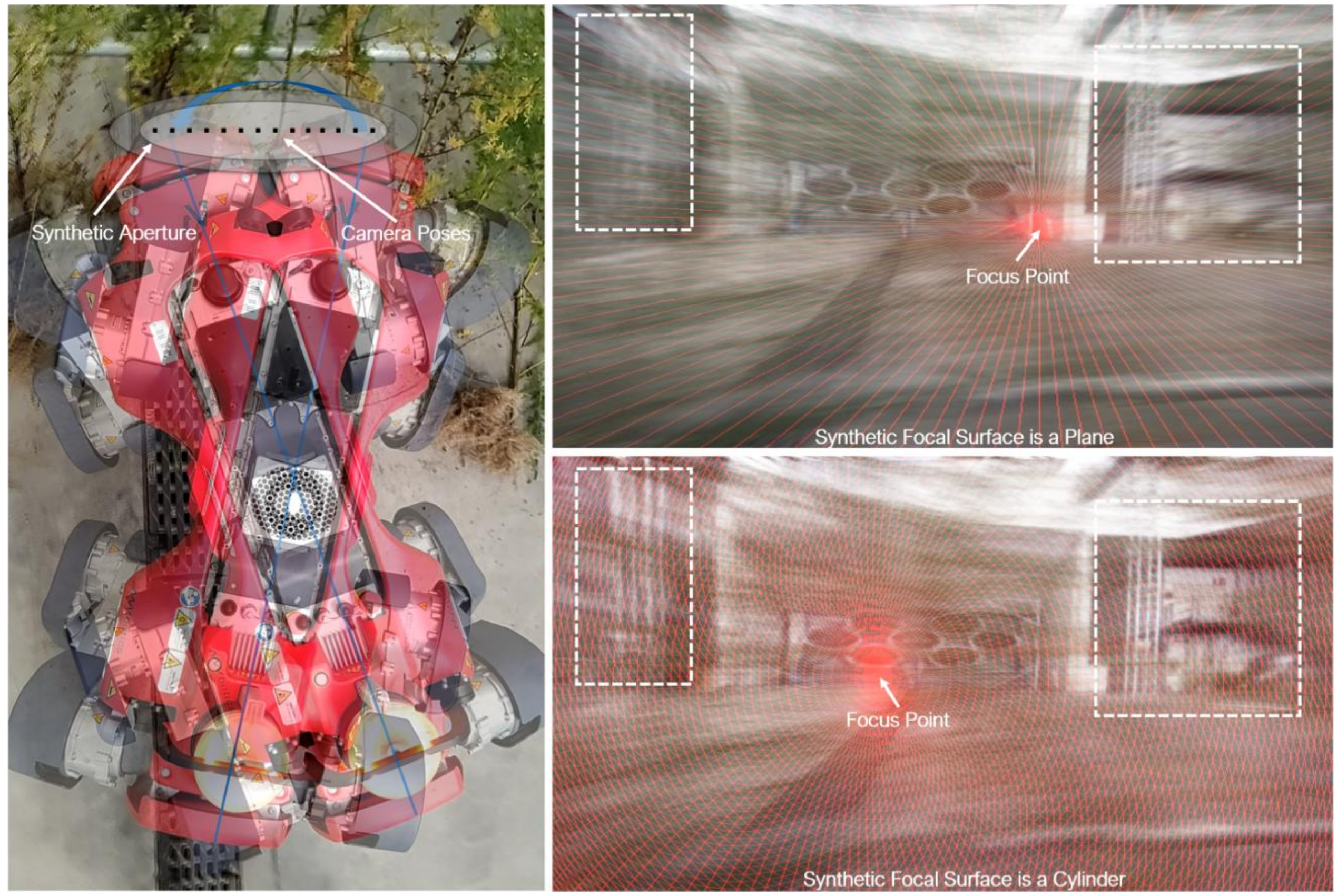

**Fig. 4. Synthetic aperture sensing via peering motion.** A peering motion (such as the axial rotation shown here) generates a synthetic aperture (left). During this motion, hundreds of images are captured at precisely measured poses. These images are then projected and averaged onto a digitally adjustable focal surface. This process produces a shallow depth-of-field integral image in which occlusions are suppressed (right). The focal surface (overlaid in red) can be parameterized to conform to a background region of interest. Because this region is non-planar in the example above, a cylindrical focal surface is more effective than a planar one, as seen when comparing the off-plane regions within the dashed boxes.

We employ a parametric focal surface that can be manipulated in all three dimensions (scaled, translated, and rotated) and can take on various shapes, such as planes, cylinders, and spheres. In our experiments, focusing is achieved manually by first selecting a focus point (distance and position) in the viewing direction of the robot, and then transforming and warping the focal surface relative to it. These focus changes are displayed in real-time. Alternatively, the focal surface parameters can be determined automatically using image-based metrics (77), similar to a conventional autofocus system. The examples shown in Fig. 4 compare a planar and a cylindrical synthetic focal surface. The cylindrical surface provides superior focus on scene elements located at varying distances. The remaining focus blur is caused by objects that are not precisely on the focal surface, combined with inaccuracies in pose estimation.

**Exploring different peering motions**

Animal peering is primarily used to enhance scene understanding with a limited visual system. To estimate depth from motion parallax using compound eyes supporting limited stereo vision , a horizontal peering motion is sufficient. Applying this bio-inspired analogy to robots for occlusion suppression also leads to superior scene comprehension, but requires slightly more sophisticated peering patterns.

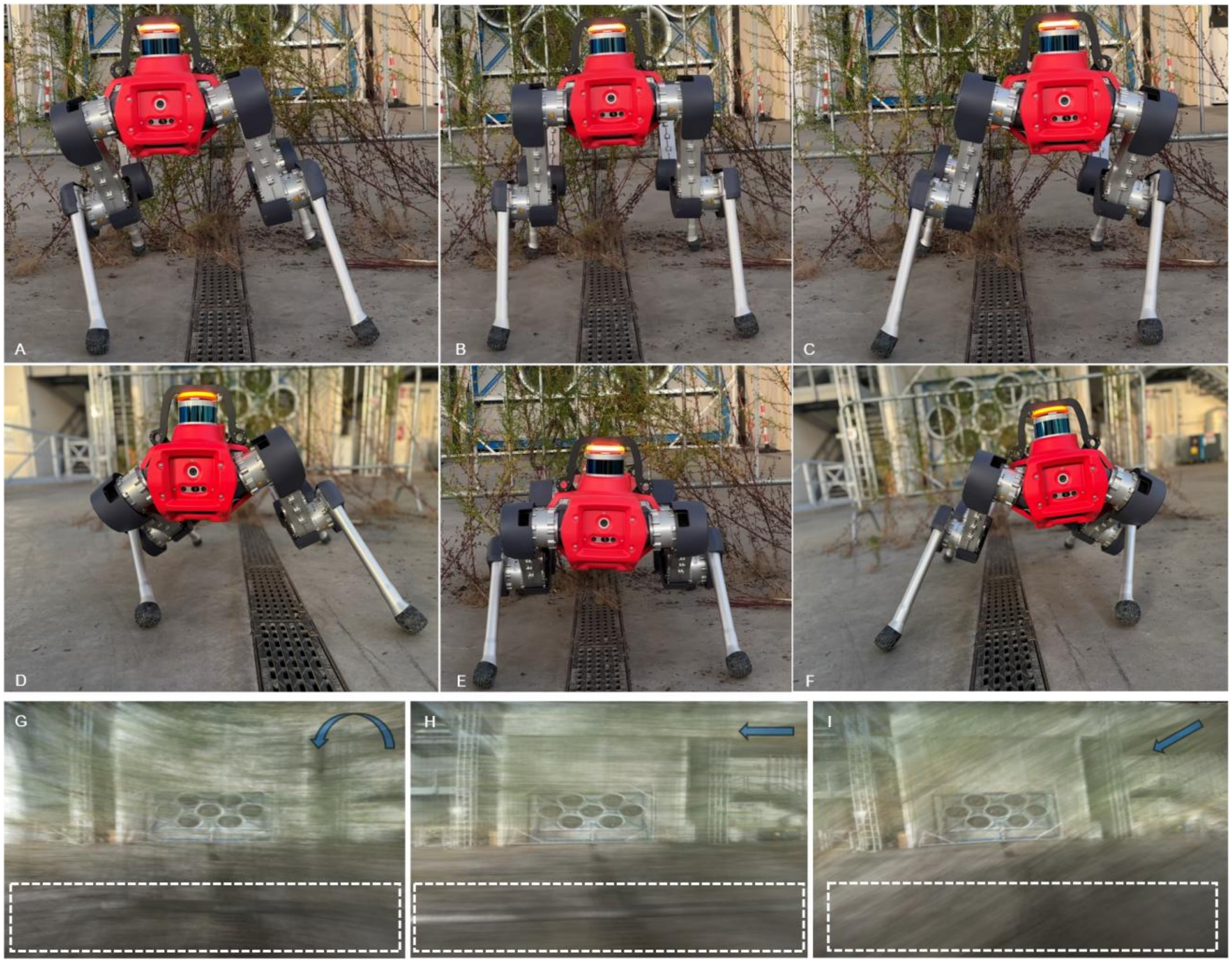

**Fig. 5. Peering as lateral shift.** Peering motions can also involve lateral shifts (A-F), as opposed to the axial rotation shown in Fig. 1. Lateral translation maximizes motion parallax, which improves occlusion removal. However, peering parallel to an occluding structure is less effective than scanning orthogonal to it. This is evident in H, where a horizontal shift leaves a horizontal fence bar visible. The results for axial rotation (G), horizontal shift (H), and diagonal shift (I) are compared, with arrows indicating the direction of motion. See Movie S2 in the *Supplementary Materials.*

Synthetic aperture sensing with a horizontal sampling pattern, for instance, struggles to suppress large horizontal occluders, as they obscure the background from many viewpoints. Our experiments (Fig. 5) show that lateral shift motions yield superior results to axial rotations, as they maximize motion parallax. Furthermore, motions with a vertical component can also suppress horizontally aligned occluders.

For peering motions that involve lateral shifts, the camera translates horizontally or vertically while remaining on the same plane, resulting in a planar or linear rather than in a circular or spherical sampling pattern. This configuration maximizes motion parallax. Note that insects, such as locusts, follow the same objective, compensating for abdomen rotation with an opposite head rotation (3). However, since rotating a robot's head is not always feasible, a lateral body shift can be used instead. The range of these shifts is constrained by the robot's mechanical limits (for the ANYbotics ANYmal, this was 30cm horizontally and 20cm vertically, Fig 5A-F). We compare three distinct peering motions (Fig. 5, bottom row): a rotational motion (Fig. 5G), as shown in Fig. 1 for reference, using a 15cm SA with 26 integrated images; a horizontal shift (Fig. 5H) with a 15 cm SA and 20 images; and a diagonal shift (Fig. 5I) with a 16cm SA and 28 images. The horizontal shift yields superior results to the rotation by maximizing parallax. Furthermore, unlike the purely horizontal motion, the diagonal shift can also suppress horizontally aligned occluders better (such as the fence's metal bar in the dotted box) because it benefits from both horizontal and vertical motion parallax. These examples also demonstrate that effective occlusion removal is achievable with an order of magnitude fewer samples, supporting the use of fast motions even with limited camera frame rates.

**Improving vision reasoning**

Partial occlusion has been shown to hinder scene understanding in basic computer vision tasks, such as classification, while performance is restored after occlusion removal (67,68). We empirically demonstrate that even the sophisticated visual reasoning capabilities of large multimodal models (LMMs) are unable to interpret occluded backgrounds properly in conventional images, but show successful understanding when presented with SA integral images where occlusion has been suppressed.

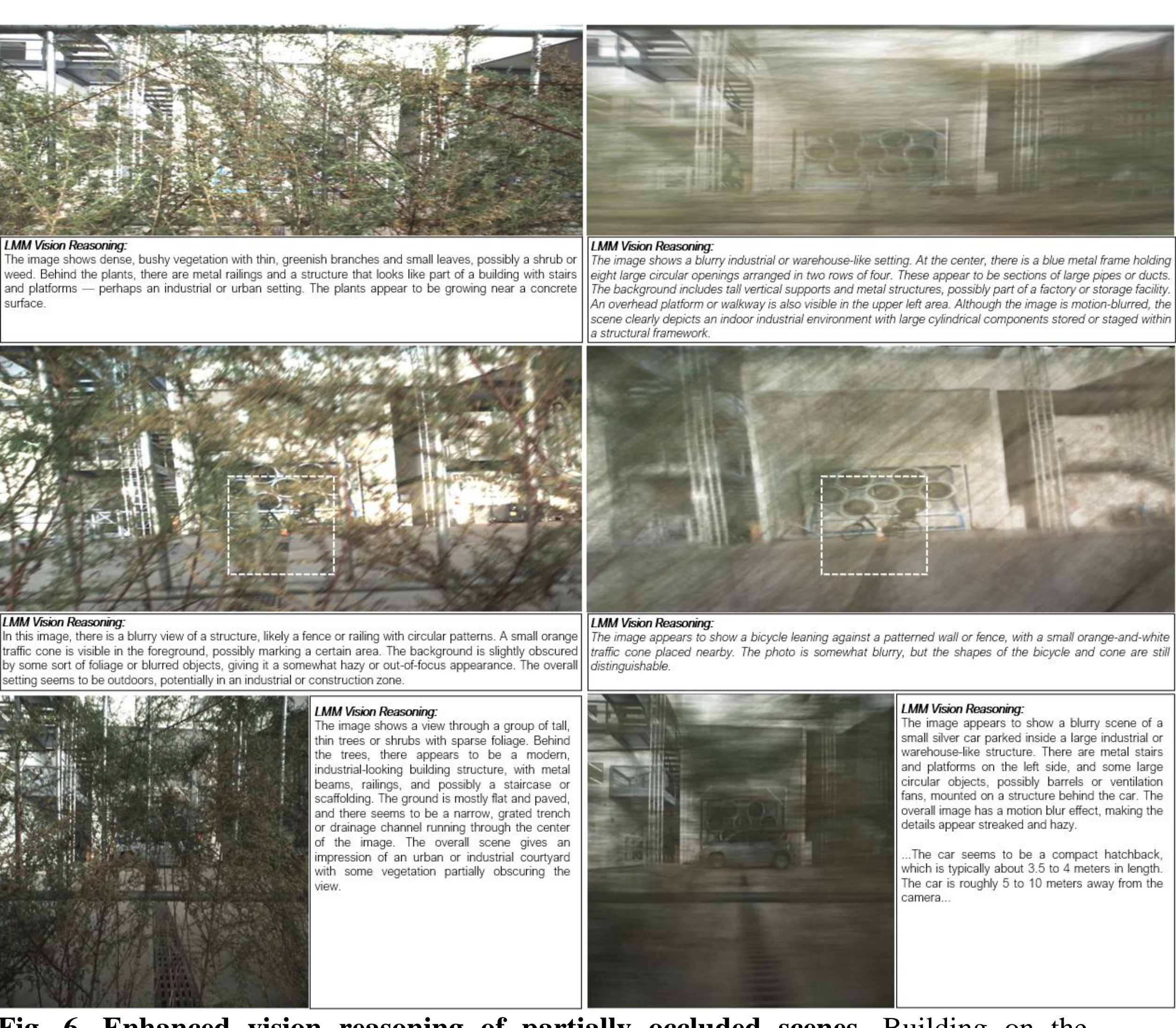

**Fig. 6. Enhanced vision reasoning of partially occluded scenes.** Building on the experiment in Fig. 1, the configurations shown here introduce greater occlusion and additional background objects. Each row displays a conventional image with its reasoning result (left) alongside the corresponding SA integral image with its reasoning result (right). Visual reasoning in conventional images focuses predominantly on foreground occluders, rendering the background uninterpretable. In contrast, reasoning improves significantly when occlusion is suppressed in the SA integral images.

In addition to results presented in Fig. 1, Fig. 6 provides the results of three additional configurations. We applied ChatGPT-5.0's vision reasoning capability for all our experiments. In our first configuration (Fig. 6, top row) we first increased the occlusion density compared to the corresponding experiment shown in Fig. 1, without changing the background scenery. We sampled 82 images with a horizontal shift peering motion over 12cm. The next configuration (Fig. 6, center row) employed a diagonal shift peering motion, capturing 32 images over 19cm, with vision reasoning applied on the region of interest (dashed box). Additional objects (a bicycle and a traffic cone) were placed in the background. The last configuration (Fig. 6, bottom row) shows the results from another horizontal shift peering motion (151 images over 9cm). In this case, a car was positioned approximately 5 meters from the robot. For the examples above and in Fig. 1, we used the prompts "*What is visible in the image?*" and (for the last configuration in addition) "*How far away is the car?*".
A similar limitation observed in 3D vision methods applies here: visual reasoning on conventional images predominantly focuses on foreground occluders, leaving the background scenery inadequately interpreted. Note that vision-based reasoning models are inherently non-deterministic. While querying the same image content multiple times yields slightly varying responses, the underlying observed pattern remains consistent. It should also be noted that in our experiments, ChatGPT's reasoning time was approximately 10-20s, varying with server load.

**The impact of occlusion masking**
The simple averaging of all overlapping pixels on the synthetic focal surface introduces residual defocus blur from occluders. To suppress this blur and achieve higher contrast and better color reconstruction we can employ occlusion masking (78), but at the cost of a higher computational time. This technique identifies potential occluders in all the input images and averages only the non-occluded pixels during synthetic aperture integration. The way occlusion masks are generated can vary. For example, they can be determined based on reflectance or depth cues.

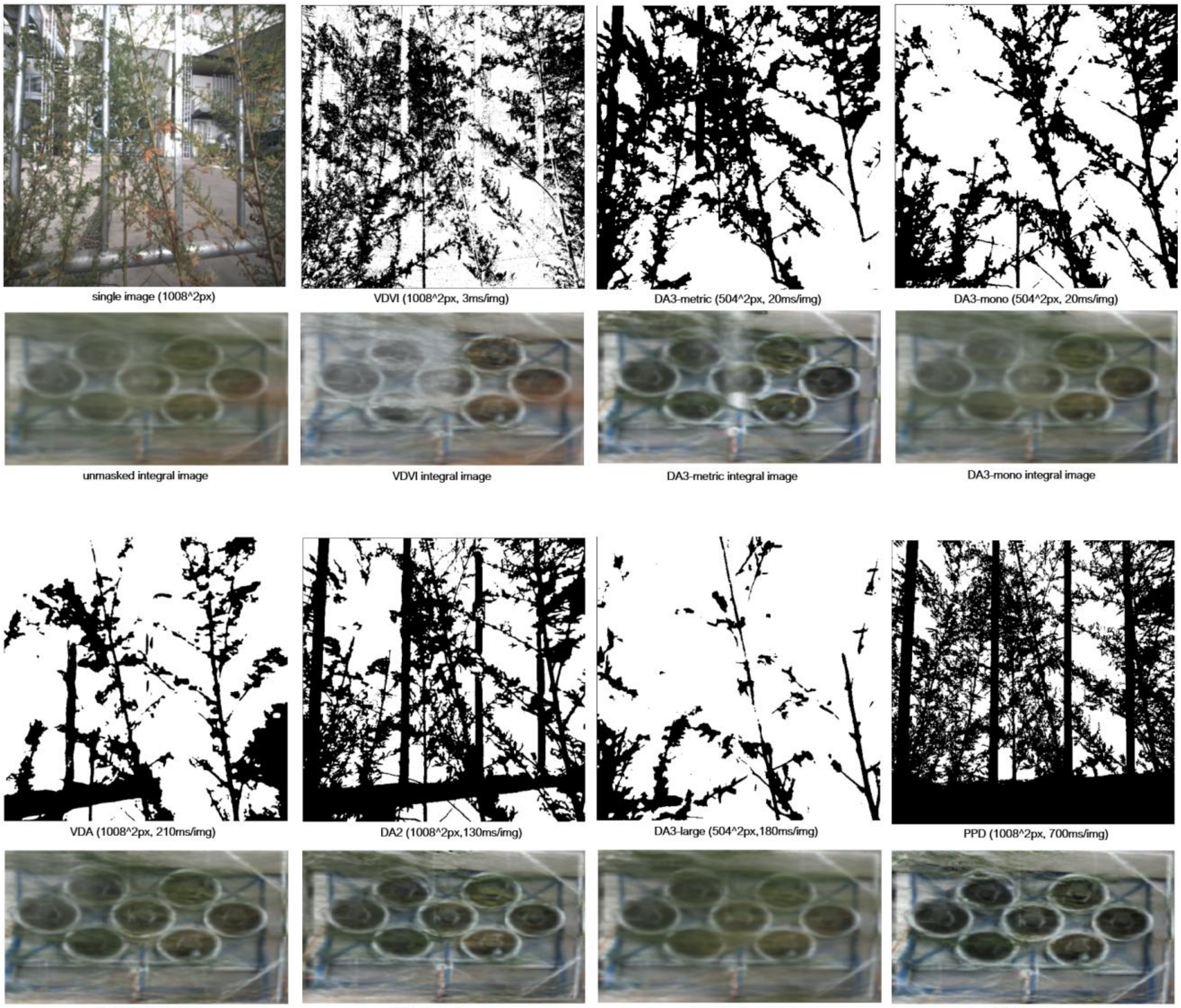

**Fig. 7. Comparing various input cues for occlusion masking and their impact on SA integrals.** The examples present identical perspectives of the unmasked input images alongside the SA integrals processed with occlusion masks derived from vegetation index (*VDVI*) and depth cues (*DA3-metric*, *DA3-mono*, *DA3-large*, *VDA*, *DA2*, and *PPD*). In the generated binary masks, zero-value pixels represent occluders, while non-zero values indicate non-occluders. The corresponding image resolutions and per-image computational times are provided. *PPD* delivers superior results but requires the longest computational time.

Figure 7 compares various image cues for occlusion masking. For example, if occluders are assumed to be plants, the Visible Difference Vegetation Index (*VDVI*) (79) can be applied to RGB images to identify and mask likely vegetation. While vegetation indices are more accurately estimated with multispectral cameras using bands beyond visible light, non-vegetative elements like the foreground metallic fence remain unmasked, preserving their defocus blur. Although 3D vision cannot reconstruct occluded backgrounds, it effectively estimates the depth of foreground occluders. These depth maps, based on the principle that occluders are closer than the background, can also be used for masking. Beyond *VDVI*, Fig. 7 evaluates six state-of-the-art, deep-learning-based depth reconstruction methods: Depth Anything v3, *DA3* (76) in three models (metric, mono, large), Video Depth Anything, *VDA* (80), Depth Anything v2, *DA2* (81), and Pixel-Perfect Depth, *PPD* (82). *DA3-large* and *VDA* are multi-view methods, while *DA3-metri*c, *DA3-mono*, *DA*2, and *PPD* are single-view. Note that the models support fixed resolutions only, and require downscaling from the original input resolution (see *Materials and Methods*). We generate binary occlusion masks for each input image by normalizing and thresholding the reconstructed depth values and vegetation indices. Figure S1 in the *Supplementary Materials* shows the occlusion masking results for a NIR recording. Deep-learning-based 3D techniques underperform on this grayscale content relative to color content, most likely due to a training data imbalance skewed toward color. Additionally, vegetation index computations are infeasible.

## MATERIALS AND METHODS

All experiments were conducted using the ANYbotics ANYmal quadruped robot. The robot is equipped with multiple vision sensors, including two FLIR Blackfly wide-angle RGB cameras mounted on the front and the rear of the robot's torso. In addition, six Intel RealSense depth cameras are installed around the robot's body, with one camera on each side and redundant sensors at the front and rear.

To execute the robot motions used during data collection, we employ native controllers provided by ANYbotics. All controllers run on the robot's onboard locomotion PC with a control loop frequency of 400Hz. Rotations around the vertical axis are achieved with a torso controller, where the desired yaw angle is specified via a joystick input. The full angular range allowed by the controller is exploited to generate the rotational motion. For horizontal and diagonal shifts of the robot's base, we use another stock controller based on Free Gait (83). This controller is a task-oriented motion framework that allows the specification of high-level task-space commands, such as desired base or end-effector motions, which are updated at 10Hz. These tasks are then tracked by a whole-body controller that computes the joint-level commands required to realize the desired motion. For our experiments, the robot's legs are first positioned in an approximately rectangular stance of 80cm x 50cm. Subsequently, the target positions of the robot's base in 3D space are specified through a configuration file using the Free Gait API, while keeping the base orientation fixed. The duration and smoothness of the motion are controlled by adjusting the weight associated with the base's velocity task. Once the motion parameters are defined, they are uploaded to the robot's onboard PC, and the desired motion is triggered via the Operator GUI.

During robot motion, we recorded RGB images from the front wide-angle camera and NIR images from one of the NIR cameras of the upper front depth sensor. Data acquisition and synchronization are handled using the Robot Operating System (ROS). Sensor data and robot state information are stored using rosbag files, which are time-synchronized recordings of ROS message streams. These rosbag files are saved locally on the robot's onboard navigation PC and later transferred to an external workstation for offline post-processing. In addition to visual data, we also record the pose of the robot's base (main body) in the world frame. Camera poses are obtained from the base poses through a fixed known rigid transformation. The robot base pose is estimated using the method described in (84), which relies on a two-state implicit filter that fuses proprioceptive joint encoder measurements with data from an onboard Inertial Measurement Unit (IMU). Since the measurements are recorded at different frequencies (images at 30fps, poses at 400Hz), we sampled and interpolated the pose data to match the image acquisition rate.

To perform synthetic aperture integration, we developed a GPU-accelerated deferred rendering application (Software S1 in the *Supplementary Materials*). Built with the CUDA toolkit, this software loads recorded images and poses either from files or via a shared memory interface. To optimize performance for GPU-based image processing, we cropped the original images to square dimensions (1080^2px for RGB, 480^2px for NIR). All computations and performance timings were conducted on a system with an Intel Core i9 processor, 64GB of RAM, and a single NVIDIA RTX 5090 GPU.

Because the deep-learning-based depth reconstruction models require fixed input resolutions (1008^2px or 504^2px), we accordingly rescaled the images to the closest-matching resolution for occlusion masking. For a fair comparison, we applied the same rescaling procedure to *VDVI*. All depth reconstruction methods were evaluated using Python 3.10 and PyTorch 2.10.0. *DA3* and *DA2* adopt a teacher–student training paradigm, where supervision is primarily derived from the DINOv2 vision foundation model (85). *VDA* extends *DA2* by replacing its original single-frame prediction head with an efficient spatiotemporal module, enabling improved temporal consistency across video sequences. *PPD* is a pixel-space diffusion transformer that introduces two key architectural innovations. First, the Semantics-Prompted Diffusion Transformer (DiT) integrates semantic representations extracted from large-scale vision foundation models to enhance fine-grained structural detail preservation and depth coherence. Second, the Cascade DiT architecture progressively refines depth predictions across multiple stages, resulting in improved reconstruction accuracy while maintaining computational efficiency.

For vision reasoning tasks, both conventional and SA integral images were provided alongside the prompts to ChatGPT 5.0. A new chat session was initiated for each image to prevent in-context learning from prior interactions.

The robot peering experiments were conducted at the Field Robotics Laboratory South Tyrol, in an 800-square-meter industrial hall and the surrounding outdoor areas. The animal peering experiments were conducted at the Department of Zoology, University of Cambridge (locust) and the School of Life Sciences (bush cricket), University of Sussex.

## DISCUSSION

Exploratory field robots have significantly expanded our ability to investigate environments that are too dangerous, remote, or inaccessible for humans. These robots enhance safety, reduce costs, and enable continuous data collection in challenging conditions. They are widely used in environmental monitoring, agriculture, mining, disaster response, space exploration, underwater research, and infrastructure inspection. By performing tasks like mapping terrain, assessing hazards, and gathering scientific data, field robots play a crucial role in advancing research, improving operational efficiency, and supporting safer decision-making across many applications. In vegetated environments, such as forests, exploratory robots play a vital role in navigating complex, cluttered environments where human access is limited and traditional equipment struggles. Equipped with advanced sensing, mapping, and mobility capabilities, these robots can move through undergrowth, around obstacles, and beneath forest canopies to gather information for biodiversity monitoring, wildfire risk assessment, conservation planning, tracking wildlife, surveillance, and studying forest health indicators such as soil conditions, canopy density, and the spread of plant diseases or parasites . Visual occlusion from obstacles like vegetation or debris at disaster sites can severely block a robot's sensors, causing conventional systems like cameras and LiDAR to yield incomplete data. This compromises accuracy and hampers scene understanding and navigation.

We show that severe occlusion creates such significant data gaps that even advanced machine learning techniques like vision reasoning fail. However, we demonstrate that "peering," a behavior insects use to overcome vision limitations, can also enable robots to significantly enhance their vision reasoning under partial occlusion.

The statistical analysis in (64) makes three key findings of synthetic aperture sensing for occlusion removal: First, its efficacy is highly dependent on the occlusion level. It performs optimally at approximately 50% occlusion. For occlusions significantly above this threshold, effective removal becomes more difficult. Conversely, for cases far below it, the application of synthetic aperture sensing is often unnecessary. Second, the size of the SA does not need to be very large (64). A minimal baseline is one that yields a disparity equal to the projected occluder size. Third, the visibility gain depends on the number of integrated samples. This number does not need to be extremely high, but must not be undersampled (64). In scenarios where occlusion masking successfully removes occluder pixels, a higher sample count increases the statistical probability of observing the background through gaps in the occluding foreground. Consequently, more samples are beneficial in such cases. However, with an adaptive SA approach (86,87) that takes more samples in sparse regions and fewer in dense ones, this number can be significantly reduced.

Our experiments were carried out on a quadrupedal robot. However, the ability to peer is not limited to such platforms, but is also applicable to bipedal, hexapod, wheeled, or crawling platforms. The peering motion can be achieved either by moving a camera payload, by moving the robot's entire frame (e.g., during navigation), or both. While sensor fusion can further enhance performance, multi-spectral and omnidirectional vision offer solutions for a wide range of applications. Adaptive sampling in response to local occlusion conditions would lead to enhanced results. All of this must be investigated in the future. For practical systems, all computations must be performed onboard within the robot's processing framework, rather than being separated into online recording and offline pre-processing as we did for experimental simplicity. Vision reasoning can be implemented either using smaller, locally deployed LMMs (such as Qwen) running directly on the robot, or via remote API calls from the robot to large-scale models, such as the latest versions of ChatGPT, Gemini, Claude, Kosmos, or LLaMA. Currently, computing SA integral images takes milliseconds on our hardware (e.g., 17ms to process 300 images at 1080^2px resolution). However, performing reasoning on these images using ChatGPT takes considerably longer, typically 10-20s per SA integral, depending on the server load. Local model deployment would significantly reduce this latency.

An alternative approach might be to apply vision reasoning directly to the raw input images, bypassing the integration step. This may seem practical when dealing with perspectives containing significant occlusion gaps. As illustrated in Fig. S2 of the *Supplementary Materials*, this approach can only detect background objects visible through sufficiently large gaps. While using multiple images may increase the number of identifiable objects, those that are heavily occluded in all views remain undetectable. Synthetic aperture integral imaging overcomes this limitation. Moreover, it enables the rapid computation of a focal stack (a series of images at varying synthetic focal distances). Applying vision reasoning to this focal stack allows object identification without prior knowledge of their distances or manual focal adjustment, as also demonstrated in Fig. S2 of the *Supplementary Materials*. Although processing image sequences increases vision reasoning times, small focal stacks (e.g., 10 layers) can be analyzed at moderate speeds of a few seconds (15-25s in our case, depending on server load).

Rather than relying on SA image integration or occlusion masking, we could instead leverage the direct image reconstruction and restoration capabilities of LMMs. The examples in *Supplementary Materials'* Fig. S3 demonstrate, however, that generating an occlusion-free image from a single or a series of conventional images introduces severe hallucinations – a consequence of the underlying occlusion itself. In contrast, removing residual blur from SA integral images significantly reduces these artifacts. This behavior aligns with vision-reasoning patterns. Given that LMMs typically employ unified architectures and shared training data for both vision reasoning and image generation, the results in Fig. S3 effectively visualize what the LMM "sees" during its reasoning process on our data. Note that like vision-based reasoning, generative AI is inherently non-deterministic.

Replacing peering with a static array of camera sensors (65) would severely undersample the SA signal. It would also result in large, complex payloads, require extremely high bandwidth to transmit multiple parallel video streams, and prevent adaptive sampling in response to local occlusion conditions. Active sensors like LiDAR and radar can also overcome occlusion but suffer from drawbacks such as low spatial resolution, a lack of multispectral data (e.g., color, temperature), and, for adequate resolutions, longer processing times for tasks like pulse compression and point-cloud registration. Peering enables real-time and high-resolution perception (which is essential for vision-based scene understanding) that is low-cost and immediately deployable on any camera-equipped robot.

The potential impact of this technology is significant. Robots that can effectively "see through" partial occlusion would possess superior perception capabilities including: scene understanding, situational awareness, and navigational capabilities in complex environments. This opens a wide range of applications. For instance, camouflage breaking is not only a subject of computer vision and image processing (88) but also a motion-parallax-based hunting strategy used by predators (89,90). This effect can also be achieved through peering, as demonstrated in simulation (see Fig. S4 in the *Supplementary Materials*).

We can learn peering strategies for robotics by studying evolved animal behaviour. This, however, creates a bidirectional exchange where robotics research and animal behaviour studies can stimulate and build upon one another. Whether peering serves functions beyond distance estimation and camouflage breaking in unoccluded environments – such as detecting camouflaged prey through occluding foliage – remains unexplored today.

## Supplementary Materials:

All supplementary materials are available at: https://doi.org/10.5281/zenodo.18093179

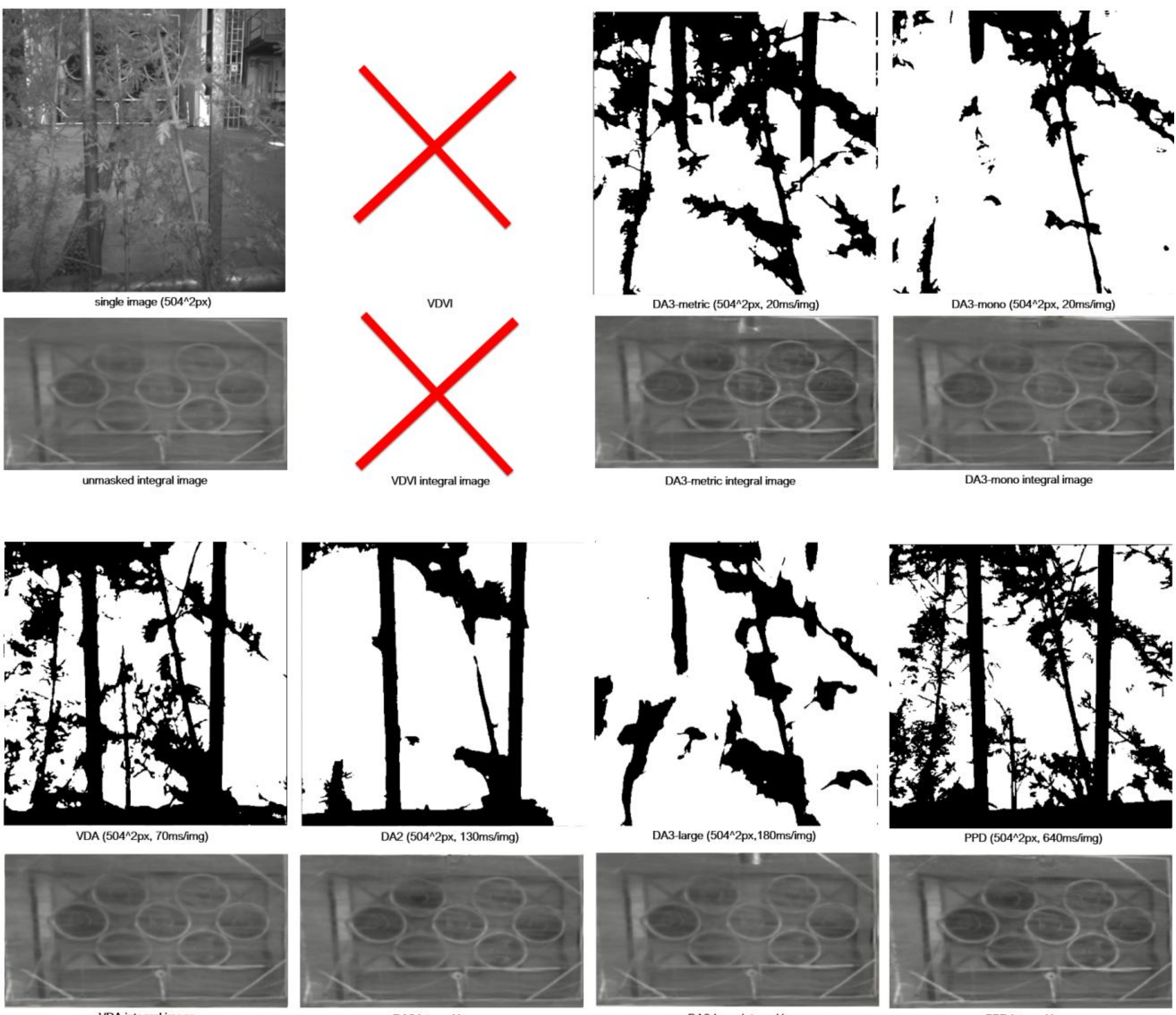

**Fig. S1. Comparing various image cues for occlusion masking and their impact on SA integrals (NIR example).** The NIR examples above were recorded simultaneously during the RGB scan in Fig. 7 of the main text. They present identical perspectives of the unmasked input images alongside the SA integrals processed with occlusion masks derived from vegetation indices (*VDVI*) and depth cues (*DA3-metric*, *DA3-mono*, *DA3-large*, *VDA*, *DA2*, and *PPD*). In the generated binary masks, zero-value pixels represent occluders, while non-zero values indicate non-occluders. The corresponding image resolutions and per-image computation times are provided. As for the RGB case, *PPD* delivers superior results but requires the most significant computation time. Here, deep-learning-based 3D vision techniques perform worse on grayscale content than on color content, while vegetation index computations are not possible.

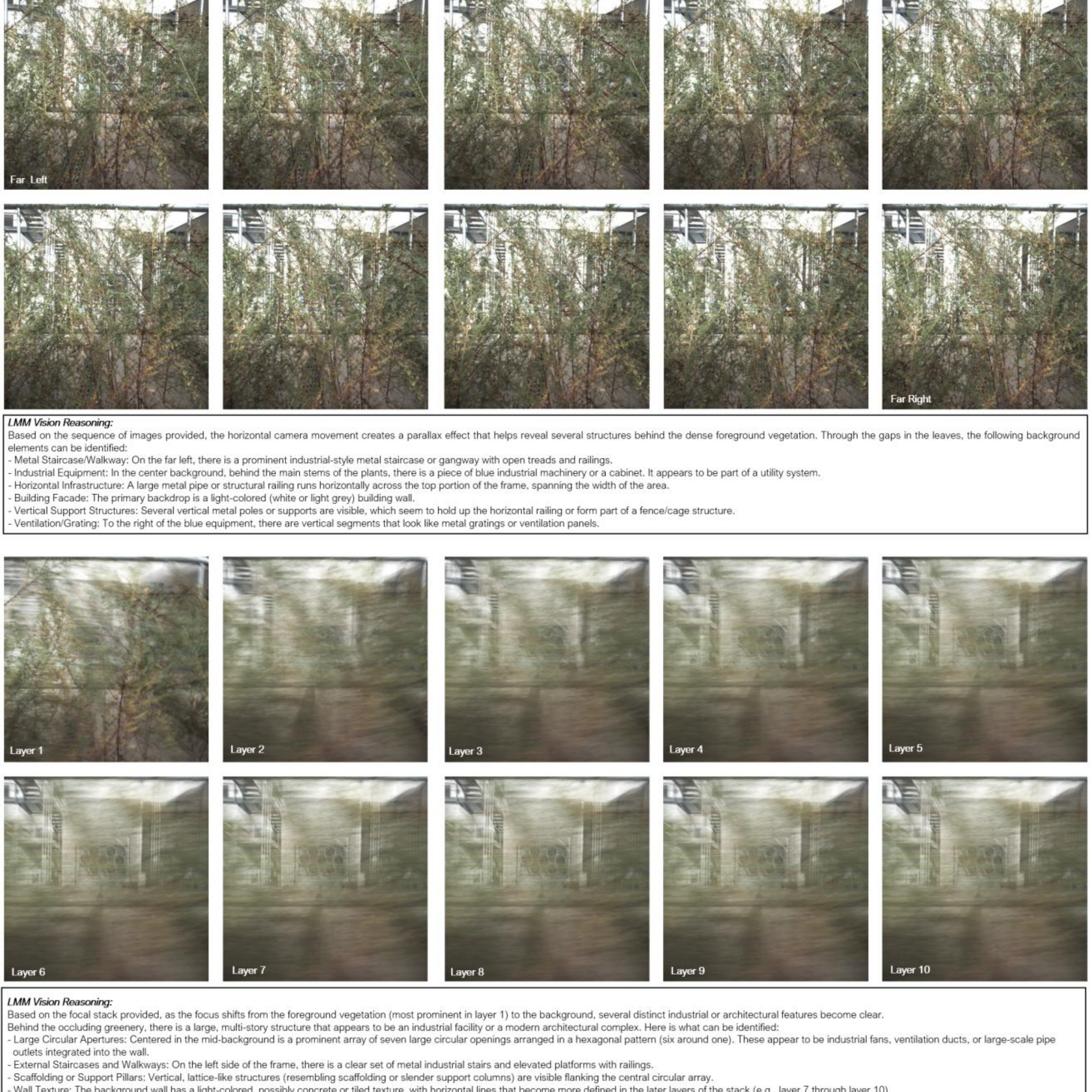

**Fig. S2. Vision reasoning on image sequences.** We demonstrate vision reasoning on two types of image sequences from the same experiment (Fig. 6, top row). Top: Vision reasoning is applied to a sequence of 10 conventional images, captured at equal distance horizontal intervals during a peering motion (from far left to far right). The prompt was: "*Here are images recorded while moving the camera horizontally. The scenery contains partially occluding vegetation in the foreground. What can be identified in the background?*" Bottom: Vision reasoning is applied to a focal stack of 10 layers (SA integral images rendered at equally spaced synthetic focal plane distances, from 3m at layer 1 to 20m at layer 10). The prompt was: "*Here is a focal stack of images. The scenery contains partially occluding vegetation in the foreground. What can be identified in the background?*". Here, the seven ventilation ducts, which remain unrecognized in the sequence of conventional images, can be identified. Note, that we used Gemini 3 for this analysis, as it supports reasoning on sequences of up to 10 images. The focal stack was computed in 170ms. Vision reasoning took around 15-25s, but depends on server load.

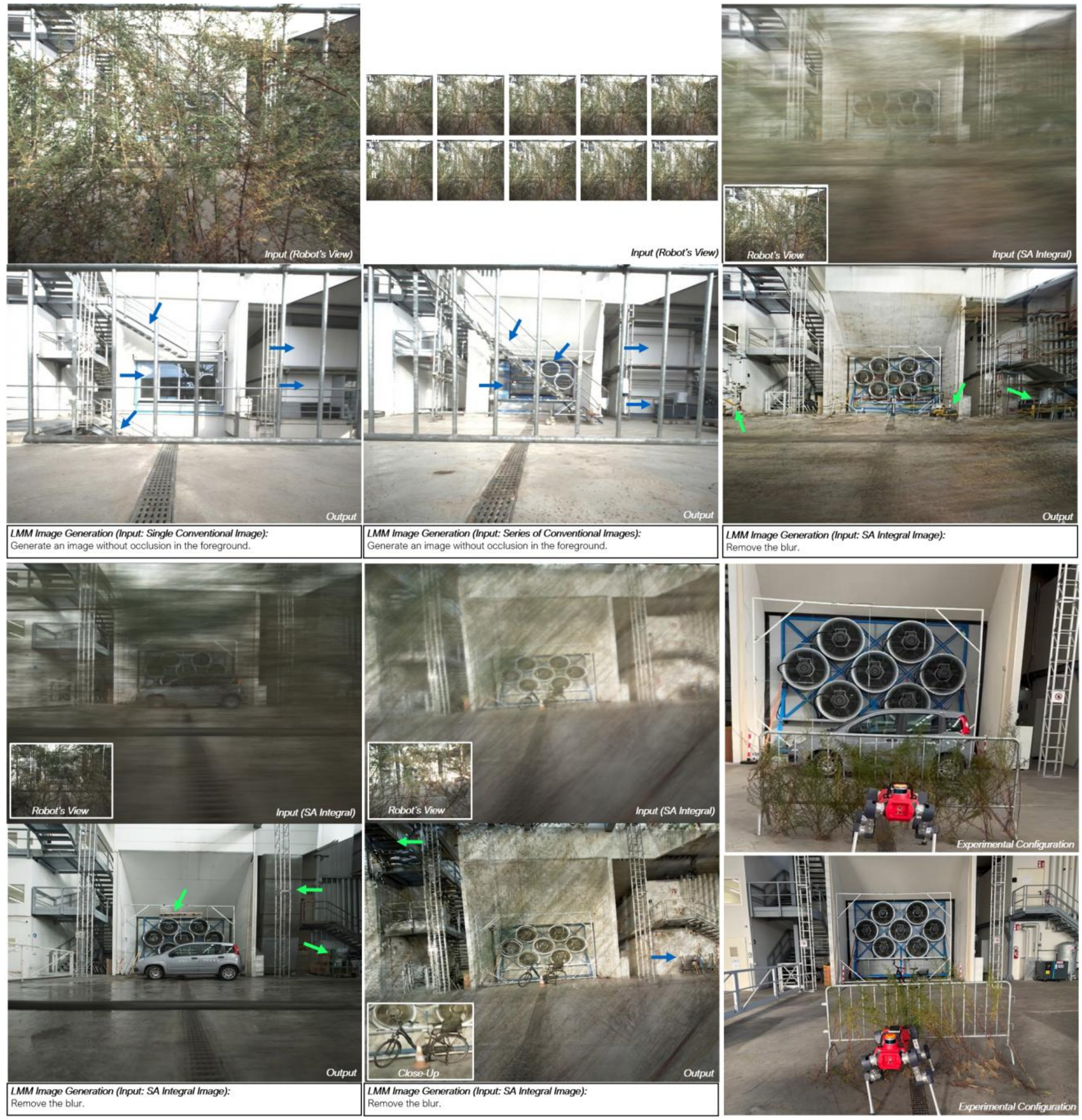

**Fig. S3. Image generation based on conventional and SA integral images.** The examples above demonstrate the LMMs' capability for image generation from various inputs: conventional images (single or in series) or SA integral images. The corresponding prompts specify the restoration task: either removing occlusions from conventional images or eliminating residual blur from SA integral images. While the outputs for conventional images exhibit significant hallucinations (though fewer for image series than for single images, as indicated by the arrows), the outputs for SA integrals show a marked reduction in such artifacts. This observed behavior aligns with the vision-reasoning results. We used Gemini 3 and its image generation model (Nano Banana).

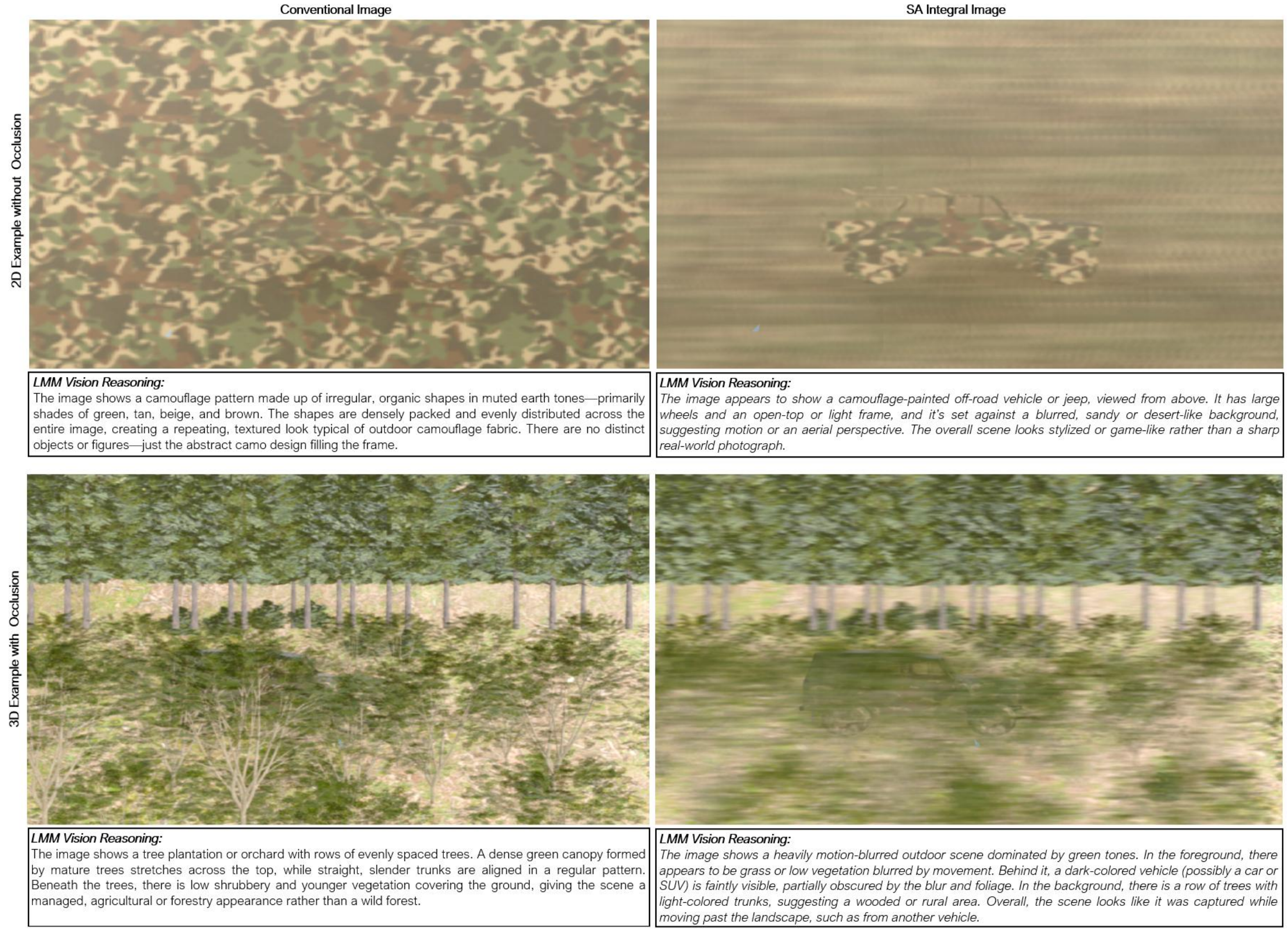

**Fig. S4. Illustrated camouflage breaking capability of peering.** Camouflaged 2D shape of an unoccluded car against a similar 2D background (top row) and a 3D graphics simulation with occlusion by vegetation (bottom row). The left column shows conventional images, while the right column shows the SA integral images focused on the car. In the 3D example, the horizontal shift peering motion used a 50cm SA with 25 samples from a distance of 6m. The 2D example also integrated 25 samples using a horizontal shift, though distances cannot be expressed in metric units. ChatGPT 5.0 responses are shown for the prompt: "What is visible in the image?".

**Movie S1. Animal and robotic peering enhance scene understanding under limited vision.** This movie supplements Fig. 1, illustrating dynamic peering motions from locusts, the Florida bush cricket, and ANYbotics' ANYmal robot. It also shows the corresponding images from the robot's perspective during peering, along with the manual focusing process within the synthetic aperture (SA) integral images (here, using a planar synthetic focal surface). Finally, it presents visual reasoning results from a large multimodal model (specifically, ChatGPT-5.0) for both indoor RGB and outdoor near-infrared (NIR) recordings.

**Movie S2. Various robotic peering motions.** This video demonstrates the peering motions implemented on a quadruped robot, including horizontal rotation, horizontal shift, and diagonal shift. These movements are constrained by the robot's equilibrium limits (avoid falling into the ground). For the ANYbotics ANYmal, this yields a maximum vertical SA of approximately 20cm and a horizontal SA of 30cm in our experiments.

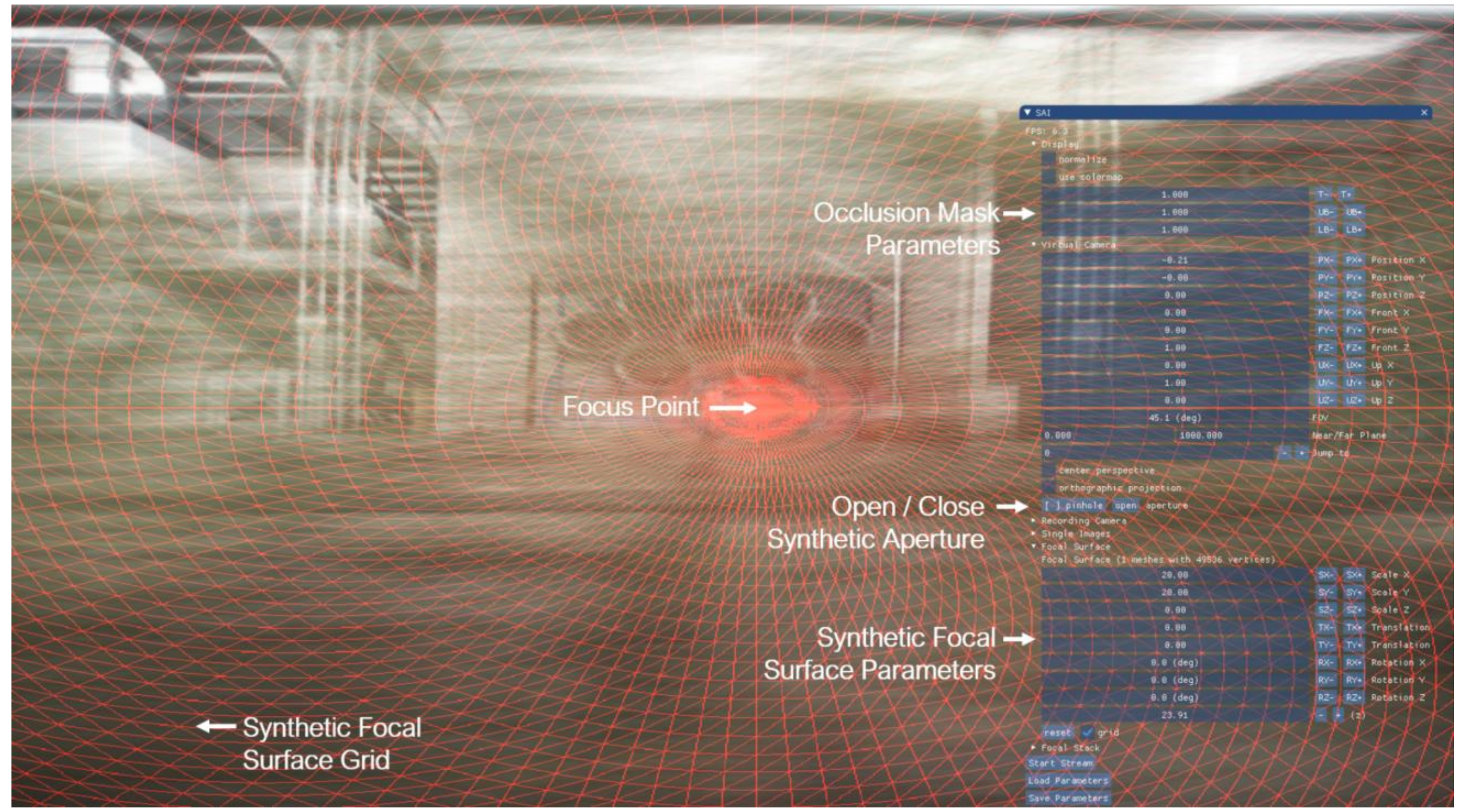

**Software S1. Synthetic Aperture Integrator.** This CUDA implementation was used to compute the SA integral images from a given set of conventional input images and the corresponding recording poses. It allows adjusting the focal surface interactively and supports occlusion masking based on Visible the Difference Vegetation Index (79). It requires Microsoft Windows and a reasonably fast GPU. Please note that the following instructions describe only the functionality for viewing the supplementary data. For more detailed information, please contact the corresponding author. To begin, choose a supplementary dataset and copy the *images* and *poses* folders, along with the *parameters.txt* file, into the main *SAI* directory. By default, the dense scene shown in Fig. S4 is preinstalled. Then, run *SAI.exe*. Please be aware that loading may take some time, depending on the number of images. Navigate the *Virtual Camera* using the mouse wheel and left button. Within the menu, you can adjust the *Focal Surface* parameters. The surface can be shifted into the scene using the *z*-parameter, where a positive value moves it forward. Additionally, it can be translated (*TX*, *TY*), rotated (*RX*, *RY*, *RZ*), and scaled (*SX*, *SY*, *SZ*). Note that the focal surface is based on a unit half-sphere. By selecting large values for *SX* and *SY*, the surface will approximate a flat plane, while a positive *SZ* value scales it in the +*z* direction. Besides planes, cylindrical and spherical focal surfaces can also be created by setting appropriate values for *SX*, *SY*, and *SZ*. The grid-flag toggles the focal surface grid visualization on or off; the center of this grid indicates the current focus point. To view the individual captured images, select the *pinhole* aperture option in the *Virtual Camera* tab and use the *Jump to +/-* buttons to browse the sequence. To return to the synthetic aperture integration view, simply click the *open* aperture option. The occlusion mask parameters (*T*,*UB*,*LB*) threshold the occlusion mask values (ranging from -1 to 1). Within this range, high values identify occluder pixels, while low values identify non-occluder pixels. The threshold *T* serves as the central value for this separation. Pixels with values recognized as occluders are assigned low alpha values, potentially becoming fully transparent (0), whereas non-occluders are assigned high alpha values, potentially becoming fully opaque (1). The *LB* and *UB* parameters define lower and upper bounds around *T* to create a smooth transition. A pixel value below *LB* is assigned an alpha of 1, and a value above *UB* is assigned an alpha of 0. For values falling between *LB* and *UB*, the alpha value is linearly interpolated between 0 and 1. This entire process is applied to each image individually, resulting in a unique alpha mask for each one. By default, these occlusion mask values are computed as the Visible Difference Vegetation Index (*VDVI*), where a high value indicates vegetation and a low value indicates non-vegetation. *VDVI* thresholds are relatively low (e.g. $T$=0.025-0.115).

**Data S1. Data for Figure 1.** This dataset contains the images, poses, and parameters for the RGB and NIR recordings used for Fig. 1. Use Software S1 to load this data (copy *images* and *poses* folders as well as *parameters.txt* to the main folder) and compute the SA integral images.

**Data S2. Data for Figure 5.** This dataset contains the images, poses, and parameters for the rotation**,** horizontal shift, and diagonal shift peering motions used for Fig. 5. Use Software S1 to load this data (copy *images* and *poses* folders as well as *parameters.txt* to the main folder) and compute the SA integral images.

**Data S3. Data for Figure 6.** This dataset contains the images, poses, and parameters for the reasoning examples used for Fig. 6. Use Software S1 to load this data (copy *images* and *poses* folders as well as *parameters.txt* to the main folder) and compute the SA integral images.

**Data S4. Data for Figures 7 and S1.** This dataset provides the non-binarized occlusion masks, images, poses, and parameters for the RGB recordings from Fig. 7 and the NIR recordings from Fig. S1. Use Software S1 to load this data: 1.) Copy *parameters.txt* and *poses* folder to the main folder. 2.) Select an occlusion mask model and copy the corresponding *masks* folder to the main folder. 3.) Copy either *images_1008* or *images_504* folder (check model resolution support in Figs. 7 and S1: either 1008^2px or 504^2px) to the main folder and rename it to *images*. 4.) Compute the SA integral images and change the occlusion mask threshold under the displays menu. Note, that if no *masks* folder is located in the main folder of the software, *VDVI* values can be computed online. If a *masks* folder is available, the mask values will be loaded from the contained .tiff files. Make sure that the mask resolution (.tiff files in the *masks* folder) matches the image resolution (.png files in the *images* folder) by selecting the images from the correct folder (either *images_1008* or *images_504*). In any case, thresholding is done with the occlusion mask parameters of the software (see Software S1).

**Data S5. Data for Figure S2.** This dataset contains the 10 conventional images and 10 focal stack layers used for Fig. S2. Use Software S1 to generate focal stacks with *Focal Stack* menu. Layers are stored in the *integrals* folder.

**Data S6. Data for Figure S4.** This dataset contains the images, poses, and parameters for the camouflage breaking examples used for Fig. S3. Use Software S1 to load this data (copy *images* and *poses* folders as well as *parameters.txt* to the main folder) and compute the SA integral images.

**Acknowledgements:** We thank Stephen M. Rogers (University of Oxford) for assistance with recording katydid peering. Katydid videos were recorded at the University of Cambridge. We thank Armando Castillo (INDICASAT AIP) for assistance recording locust peering.

**Funding:**
Austrian Science Fund (FWF)/German Research Foundation (DFG) grants I 6046-N P "Wide Synthetic Aperture Sampling for Motion Classification" and 32185-NBL "Wide Synthetic Aperture Sampling" (OB), Cohesion Italy 21-27 Project EFRE1079 "Forest Robotic Monitoring and Automation (FORMA)" CUP I53C24001750006 Cofunded by the EU and the Autonomous Province of Bolzano (KDvE).

**Author contributions:**
Conceptualization: OB
Methodology: OB
Software: RJAAN, GL, MC, MY
Investigation/Implementation: OB, KDvE, MH, RJAAN, GL, MC, MY, SMOS, JEN
Resources: KDvE, OB, JEN, MH
Visualization: OB, RJAAN, GL, MH, MY, JEN
Funding acquisition: OB, KDvE, MH
Project administration: OB, KDvE, MH
Supervision: OB, KDvE, MH
Writing – original draft: OB
Writing – review & editing: OB, KDvE, MH, GL, MY, JEN

**Competing interests:**
Authors declare that they have no competing interests.

**Data and materials availability:**
All data are available in the main text or the supplementary materials.